\documentclass[10pt,twocolumn]{article}

\usepackage[T1]{fontenc}
\usepackage[utf8]{inputenc}
\usepackage{mathptmx}

\usepackage[letterpaper,margin=0.75in,columnsep=0.28in]{geometry}
\usepackage{titlesec}
\usepackage{graphicx}
\usepackage{booktabs}
\usepackage{array}
\usepackage{caption}
\usepackage{float}
\usepackage{indentfirst}
\usepackage{amsmath}
\usepackage[hidelinks]{hyperref}

\renewcommand{\thesection}{\Roman{section}}
\renewcommand{\thesubsection}{\Alph{subsection}}

\titleformat{\section}
  {\normalfont\fontsize{10}{12}\scshape\centering}
  {\thesection.}{0.5em}{\MakeUppercase}
\titlespacing*{\section}{0pt}{12pt plus 2pt minus 2pt}{4pt}

\titleformat{\subsection}
  {\normalfont\fontsize{10}{12}\itshape}
  {\thesubsection.}{0.5em}{}
\titlespacing*{\subsection}{0pt}{8pt plus 2pt minus 2pt}{3pt}

\makeatletter
\newenvironment{numrefs}
  {\section*{\centering\normalfont\fontsize{10}{12}\scshape VIII.\ References}
   \begin{list}{[\arabic{enumi}]}{%
      \usecounter{enumi}
      \setlength{\labelwidth}{2.2em}\setlength{\leftmargin}{2.4em}
      \setlength{\labelsep}{0.3em}\setlength{\itemsep}{2pt}\setlength{\parsep}{0pt}
      \footnotesize}}
  {\end{list}}
\makeatother

\begin{document}

\twocolumn[
  \begin{center}
    {\fontsize{17}{20}\bfseries EcoBin: A Two-Stage Deep Convolutional Neural Network for Contamination-Aware Waste Classification\par}
    \vspace{0.9em}
    {\fontsize{11}{13}\selectfont Raghav Senthil Kumar\par}
    {\fontsize{10}{12}\itshape BASIS Phoenix\par}
    {\fontsize{10}{12}\selectfont Phoenix, Arizona, USA\par}
    {\fontsize{10}{12}\selectfont senthilkumaraghav@gmail.com\par}
    \vspace{1.2em}
  \end{center}
]

\noindent{\bfseries\textit{Abstract}\textbf{---}Waste classification models have become highly accurate at sorting waste, often exceeding 95\% on benchmark datasets. However, these models fail to account for contamination in recyclable waste. We present EcoBin, a two-stage deep convolutional neural network that classifies household waste by its disposal pathway and that explicitly accounts for contamination. The first stage is a base waste classifier built on an EfficientNetV2-S backbone that assigns each of the thirty waste categories in our dataset to one of four disposal pathways. The second stage is a contamination classifier that inspects any item routed toward recycling and overrides the decision to garbage when contamination is detected. Because no public dataset of contaminated recyclables exists, we synthesize one by segmenting images of clean recyclable objects with a U2-Net model and compositing realistic contamination textures onto their surfaces. The first stage achieves 87.42\% test accuracy and a 96.13\% pathway-adjusted accuracy. Meanwhile, the contamination stage distinguishes clean from contaminated items with a 0.99 ROC-AUC. On a test set of contaminated recyclables, the complete pipeline routes 24 of 25 items correctly, compared with only 1 of 25 for the base classifier alone. A McNemar's test confirms that the improvement contributed by the contamination stage is statistically significant ($p \approx 2.4\times10^{-7}$).\par}

\section{Introduction}
The United States generated more than 292 million tons of municipal solid waste in 2018 alone [1]. Moreover, according to the World Bank's \textit{What a Waste 3.0} report, the total amount of waste generated globally is set to increase 70\% by 2050 [2]. What makes these figures particularly frustrating is that nearly three quarters of our waste stream could in principle be diverted from landfills; yet the actual recycling yield rate is only 32.1\% [1]. The gap between what could be recycled and what actually is recycled is therefore the central problem that motivates this work.

We attribute that gap to two complementary failures. The first failure, which is commonly described as wish-cycling, occurs when a well-intentioned person places a non-recyclable item into a recycling bin in the hope that it will somehow be recovered. The consequence is that materials such as plastic bags and expanded polystyrene jam sorting machinery and contaminate legitimate recyclable items. The second failure, which we describe as under-recycling, occurs when a recyclable item is mistakenly discarded as garbage and therefore never reaches a recycling facility at all. Nearly 76\% of recyclable items such as paper, cardboard, and metal end up buried in a landfill because they are missorted during disposal at the household level [3]. Both failures share a common root, namely that the person making the disposal decision is uncertain about which bin a given item belongs in.

Fortunately, advances in computer vision, and in particular deep convolutional neural networks, have made it feasible to automate this task. Waste classification models are able to classify a photograph of a waste item and deliver guidance to the user regarding how it should be disposed of [4]. These models, however, share a limitation, namely that even when an item is correctly identified by its material, it may carry contamination that renders it unsuitable for recycling. A waste classifier that reasons only about the object type will route it incorrectly. Consider the example of a pizza box: a model that recognizes the box as cardboard will confidently send it to a recycling pathway, even though the grease and food residue soaked into the cardboard mean that the correct destination is in fact the garbage. A practical sorting aid must therefore reason not only about what an item is but also about the condition that it is in.

The contributions of this work are the following. First, we present a two-stage classification pipeline in which a base classifier identifies the disposal pathway of a waste item and a contamination classifier overrides recycling decisions when contamination is present. Second, we describe a procedure for synthesizing a large dataset of contaminated recyclables, which we construct by segmenting clean objects and compositing contamination textures onto them, in order to overcome the absence of any suitable public dataset. Third, we map a thirty-class taxonomy onto the four disposal pathways defined by a real municipal recycling program, which grounds the model's output in concrete disposal guidance that a resident can act upon. Fourth, we validate the contamination stage with a McNemar's test on real photographs of contaminated recyclables.

\section{Related Work}
The application of machine learning to waste classification has grown rapidly over the past decade, and a recent systematic review reports that convolutional neural networks have become the dominant approach to vision-based waste recognition, with transfer learning from large pretrained backbones serving as the most common strategy for coping with the limited size of waste datasets [4]. A considerable portion of this progress has been organized around a small number of public datasets.

We examine three such datasets in this section: TrashNet, TACO, and TrashBox. The widely used TrashNet collection groups roughly two thousand images into six material categories, namely glass, paper, metal, plastic, cardboard, and a residual trash class [5]. The TACO dataset extends the setting to litter photographed in natural environments and provides instance-level annotations across dozens of fine-grained classes [6]. Finally, the TrashBox dataset enlarges the scale considerably and introduces new categories such as electronic waste and medical waste that earlier datasets did not contain [7].

However, the most important limitation of the existing body of waste classification datasets is that they are not suitable to train a waste classification model to reason about the condition of an object, and consequently those models cannot account for contamination. Work that addresses contamination directly remains rare, although ContamiNet demonstrated that a convolutional network trained on a large corpus of annotated bins could detect contamination in residential recycling and compost at a level approaching that of human experts [8], which establishes the feasibility of contamination-aware waste classification.

Nevertheless, a practical obstacle stands in the way of training a contamination-aware classifier, namely that no public dataset of contaminated recyclables exists. Whereas clean waste imagery is abundant, paired examples of the same object in clean and contaminated conditions are essentially unavailable. We address this obstacle by synthesizing a dataset of contaminated recyclables programmatically. To our knowledge, the combination of a pathway-level base classifier with a synthetically trained contamination classifier, integrated into a single deployed pipeline, has not been reported previously.

\section{EcoBin Design}
\subsection{Waste Classification Pipeline Overview}
\begin{figure}[t]
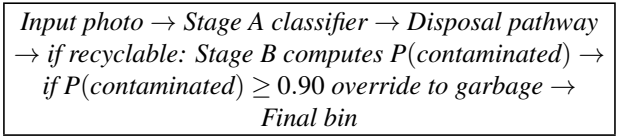

\centering
\setlength{\fboxrule}{0.6pt}
\framebox{\parbox{0.92\columnwidth}{\centering\itshape
Input photo $\rightarrow$ Stage A classifier $\rightarrow$ Disposal pathway $\rightarrow$ if recyclable: Stage B computes $P(\text{contaminated})$ $\rightarrow$ if $P(\text{contaminated}) \geq 0.90$ override to garbage $\rightarrow$ Final bin}}
\caption{The EcoBin pipeline. Stage A assigns a disposal pathway. Then, the recyclable predictions are inspected by Stage B, and the pathway is overridden to garbage when the contamination probability exceeds the threshold.}
\label{fig:pipeline}
\end{figure}

EcoBin is composed of two classifiers that operate in sequence, and we refer to them throughout as Stage A and Stage B. Stage A is the base waste classifier, and its task is to map a photograph of a waste item to one of four disposal pathways, namely curbside recycling, drop-off recycling, compost, or garbage. Stage B is the contamination classifier, and its task is to inspect any item that Stage A has routed toward a recycling pathway (curbside recycling or drop-off recycling) and to determine whether that item is clean enough to be recycled or whether it is contaminated and it should be redirected to the garbage. Figure~\ref{fig:pipeline} summarizes how the two stages cooperate.

The two stages are joined by a deterministic override rule to re-route contaminated recyclables to the garbage bin. Critically, Stage A produces a predicted class which we map to a pathway, and if that pathway is either curbside recycling or drop-off recycling then we consult Stage B for the probability that the item is contaminated, which we compute as one minus the probability assigned to the clean class. When that contamination probability meets or exceeds a threshold, we override the pathway to garbage, and otherwise we retain the recycling decision. We set the threshold to a high value of 0.90 so that the model doesn't mistakenly route clean recyclables to garbage.

\subsection{Transfer Learning and EfficientNetV2-S Backbone}
Both stages of EcoBin are built on transfer learning, which is the practice of beginning from a network whose weights have already been trained on a large general-purpose dataset and then adapting that network to a narrower task [9]. For EcoBin, we used transfer learning from ImageNet. The motivation for this choice is that a network trained from scratch on a few thousand waste photographs would have to learn elementary visual primitives such as edges, colors, and textures entirely from the limited data available, whereas a network that has already been trained on the ImageNet collection of more than a million natural images has learned those primitives already and can devote its limited waste-specific data to learning waste-specific patterns [10]. The literature on waste classification reflects this reasoning, since the most successful systems overwhelmingly fine-tune pretrained backbones rather than training from scratch [4].

We selected EfficientNetV2-S as the backbone for both stages, because it offers a favorable balance between accuracy and computational cost, having been designed to train faster and with fewer parameters than earlier networks of comparable accuracy [11]. A further practical advantage of EfficientNetV2-S is that it incorporates input rescaling and normalization within the network itself, which means that raw pixel values can be supplied to the model without a separate preprocessing step, and this simplifies both training and the eventual deployment of the model behind a web interface.

\subsection{Data Augmentation}
We use data augmentation techniques to discourage the model from memorizing incidental properties of the training images, such as a particular orientation or a particular lighting condition, rather than learning the genuine visual structure that distinguishes one class from another [12]. Augmentation applies a randomized set of transformations to each training image on every epoch, so that the network effectively sees a slightly different version of each example each time, and this broadens the apparent diversity of the training set without any additional data collection. Table~\ref{tab:aug} lists the six augmentation operations that we apply, together with the effect of each. We use the same augmentation pipeline for both stages.

\begin{table}[t]
\centering
\footnotesize
\begin{tabular}{@{}p{0.42\columnwidth}p{0.46\columnwidth}@{}}
\toprule
\textbf{Operation} & \textbf{Effect} \\
\midrule
RandomRotation(30/360) & Rotates the image by up to $\pm30$ degrees \\
RandomZoom(0.2) & Zooms in or out by up to 20 percent \\
RandomTranslation(0.15, 0.15) & Shifts the image by up to 15 percent \\
RandomFlip (horizontal and vertical) & Mirrors the image horizontally and vertically \\
RandomBrightness(0.15) & Adjusts brightness by up to 15 percent \\
RandomContrast(0.15) & Adjusts contrast by up to 15 percent \\
\bottomrule
\end{tabular}
\caption{The six data-augmentation operations applied during training. The same pipeline is shared by Stage A and Stage B and is disabled at evaluation.}
\label{tab:aug}
\end{table}

\subsection{Two-Phase Training Strategy}
We train both stages with the same two-phase procedure, and the purpose of dividing training into two phases is to adapt the pretrained backbone to waste imagery without destroying the general features that make transfer learning worthwhile in the first place. In the first phase we freeze the backbone entirely and train only the new classification head, so that the randomly initialized head can settle into a reasonable configuration without disturbing the pretrained weights. This phase is fast because almost all of the network's parameters are held fixed. In the second phase we unfreeze a portion of the backbone, namely the top half for Stage A and the top third for Stage B, and we allow those layers to adapt to waste imagery using a small learning rate.

A risk arises during the second phase that, if neglected, produces a sharp and characteristic drop in accuracy at the moment of unfreezing, and we address it by keeping every batch-normalization layer in inference mode throughout fine-tuning. Batch-normalization layers maintain running estimates of the mean and variance of their inputs, and when a frozen backbone is suddenly made trainable those estimates begin to drift on the small fine-tuning batches, which disrupts the very feature statistics that the head was trained against [13]. By freezing the batch-normalization layers we keep the backbone behaving during fine-tuning exactly as it behaved during the head warmup, and as a result the accuracy curve climbs smoothly through the transition rather than dipping.

We pair this strategy with a learning-rate schedule that warms up linearly to a small peak and then decays along a cosine curve to a near-zero floor, a shape that is closely related to the cosine annealing schedules used elsewhere in deep learning [14]. The warmup gives the newly unfrozen gradients a few hundred steps to settle before the learning rate reaches its peak, and the cosine decay then cools the rate gradually so that the final epochs make only fine adjustments. We optimize with the Adam algorithm [15].

\section{Stage A: Base Waste Classifier}
\subsection{Dataset and Disposal-Pathway Mapping}
To train Stage A we use the Recyclable and Household Waste Classification dataset, which provides more than fifteen thousand images distributed evenly across thirty classes, with five hundred images per class [16]. The classes span a broad range of household waste items including plastic, paper, cardboard, glass, metal, organic waste and textiles. A particularly valuable property of this dataset for our purposes is that each class is divided into two subsets: a default subset that depicts the item in a clean, studio-style setting and a real-world subset that depicts the item photographed in an ordinary environment.

\begin{figure}[t]
\centering
\includegraphics[width=0.98\columnwidth]{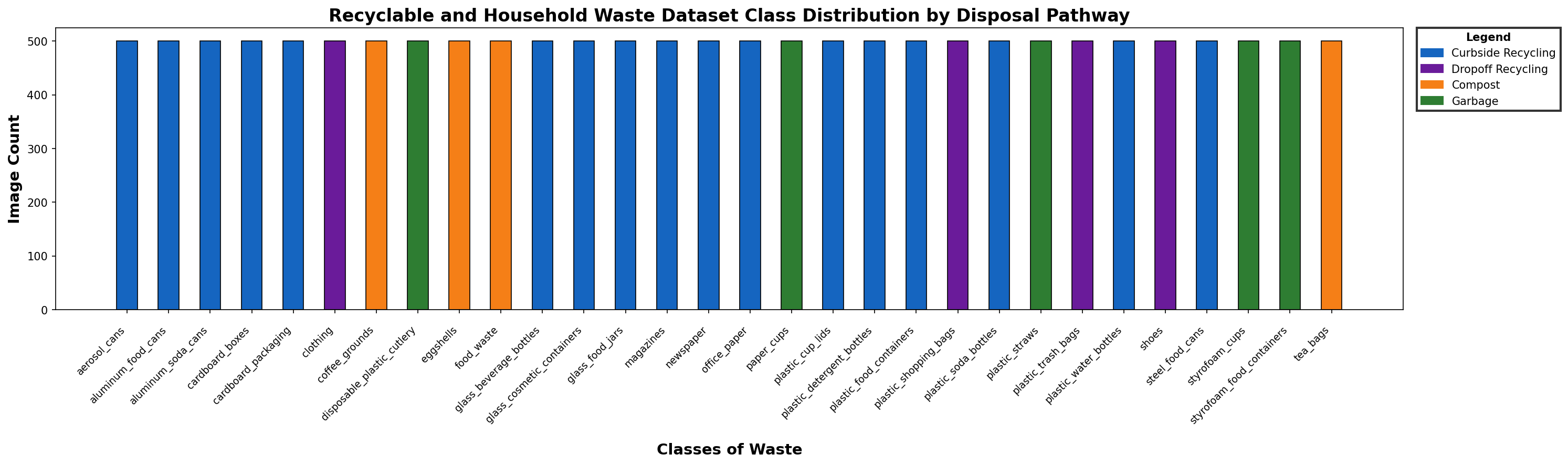}
\caption{Class distribution of the Recyclable and Household Waste Classification dataset. All thirty classes are balanced at five hundred images each, and each bar is colored by the disposal pathway to which the class is mapped.}
\label{fig:classdist}
\end{figure}

The thirty object classes are not, however, the information that a user needs. Therefore, we map each class onto one of four disposal pathways. Because recycling rules are determined locally and vary from one municipality to another, we adopt the guidelines of a single concrete program, namely the recycling rules of the City of Phoenix. We note that a hazardous-waste pathway exists in principle, but no class in the dataset falls under it, and so we exclude it. Figure~\ref{fig:classdist} shows each waste class in the dataset mapped to its appropriate disposal pathway.

\subsection{Train, Validation, and Test Split}
We partition the data into a training set of seventy percent, a validation set of fifteen percent, and a test set of fifteen percent, which yields 10{,}500 training images, 2{,}250 validation images, and 2{,}250 test images. We further divide the test set into a default subset of 1{,}140 images and a real-world subset of 1{,}110 images, and this division allows us to quantify the domain gap between studio-style imagery and ordinary photography directly. Additionally, we fix a random seed for every split and shuffle so that the partitioning is reproducible and a reader who re-runs the procedure obtains the same division of data.

\subsection{Architecture}
The Stage A architecture attaches a compact classification head to the EfficientNetV2-S backbone. The feature map produced by the backbone is reduced to a single vector by global average pooling, after which a batch-normalization layer stabilizes the activations and a dropout layer randomly zeroes a fraction of them in order to discourage co-adaptation among features [17]. A dense layer of 256 units with an L2 weight penalty then learns the waste-specific representations. A second dropout layer provides further regularization, and a final dense layer produces a probability distribution over the thirty classes.

\begin{table}[t]
\centering
\footnotesize
\begin{tabular}{@{}lll@{}}
\toprule
\textbf{Layer} & \textbf{Output shape} & \textbf{Params} \\
\midrule
Input & (224, 224, 3) & 0 \\
Augmentation & (224, 224, 3) & 0 \\
EfficientNetV2-S & (7, 7, 1280) & 0 \\
GlobalAvgPool2D & (1280) & 0 \\
BatchNorm & (1280) & 5{,}120 \\
Dropout(0.3) & (1280) & 0 \\
Dense(256, Swish) & (256) & 327{,}936 \\
Dropout(0.2) & (256) & 0 \\
Dense(30, Softmax) & (30) & 7{,}710 \\
\bottomrule
\end{tabular}
\caption{Stage A architecture with output shape and parameter count for each layer.}
\label{tab:archA}
\end{table}

\section{Stage B: Contamination Classifier}
\subsection{Synthetic Recyclable Contamination Dataset}
The central obstacle to training Stage B is that no public dataset of contaminated recyclables exists, so we had to construct one synthetically. The guiding idea is straightforward: take a clean recyclable photograph from the Stage A dataset, segment the object within it, paste randomized contamination textures onto the object's surface, and save the contaminated result alongside the unmodified clean original. Each source image yields one clean copy and eight contaminated copies, one for each of eight contaminant types. We restrict synthesis to the recyclable classes, because contamination is only a deciding factor for those items.

Segmentation is the step that determines where contamination may legitimately be placed, and we perform it with a U2-Net model accessed through the rembg library [18]. We run the segmentation model on the source image and it returns a soft mask of the object, after which a morphological closing operation fills small interior holes that arise from transparent or reflective surfaces such as glass, and a Gaussian blur feathers the edges of the mask so that contamination placed near the object boundary fades naturally rather than cutting off in a hard line.

Given a clean object and its mask, we composite contamination onto the object. For each contaminated copy we choose a random anchor point that lies within the object region. The texture is then alpha-blended into the underlying image using a three-way weight, formed from the texture's own transparency, the object mask, and an opacity scalar. This prevents the contamination from spilling outside the object boundary. We draw each contaminated copy's severity uniformly from light, medium, and heavy levels, where heavier severities correspond to a larger number of pasted patches and therefore to more visible contamination.

We applied this procedure to one thousand source images randomly sampled across the recyclable classes in proportion to their size, and because each source yields nine outputs the complete synthetic dataset contains nine thousand images.

\subsection{Architecture}
Stage B reuses the trained Stage A backbone and replaces only the final classification head, so that it inherits all of the visual features that Stage A learned and need only learn to distinguish clean items from contaminated ones. The new head has the same structure as the Stage A head, differing only in its output layer, which produces a nine-way softmax over the clean class and the eight contaminant types rather than the thirty-way softmax of Stage A.

\subsection{Training}
Stage B training mirrors the two-phase procedure of Stage A, with a first phase that trains only the new head on the frozen backbone and a second phase that unfreezes the top third of the backbone under a warmup-cosine schedule with a small peak learning rate. Additionally, we weight the loss by class in order to compensate for the deliberate one-to-eight imbalance between clean and contaminated images, so that the abundance of contaminated examples does not bias the model against ever predicting the clean class.

\section{Experimental Results Evaluation}
\subsection{Stage A Results}
\begin{figure}[t]
\centering
\includegraphics[width=0.98\columnwidth]{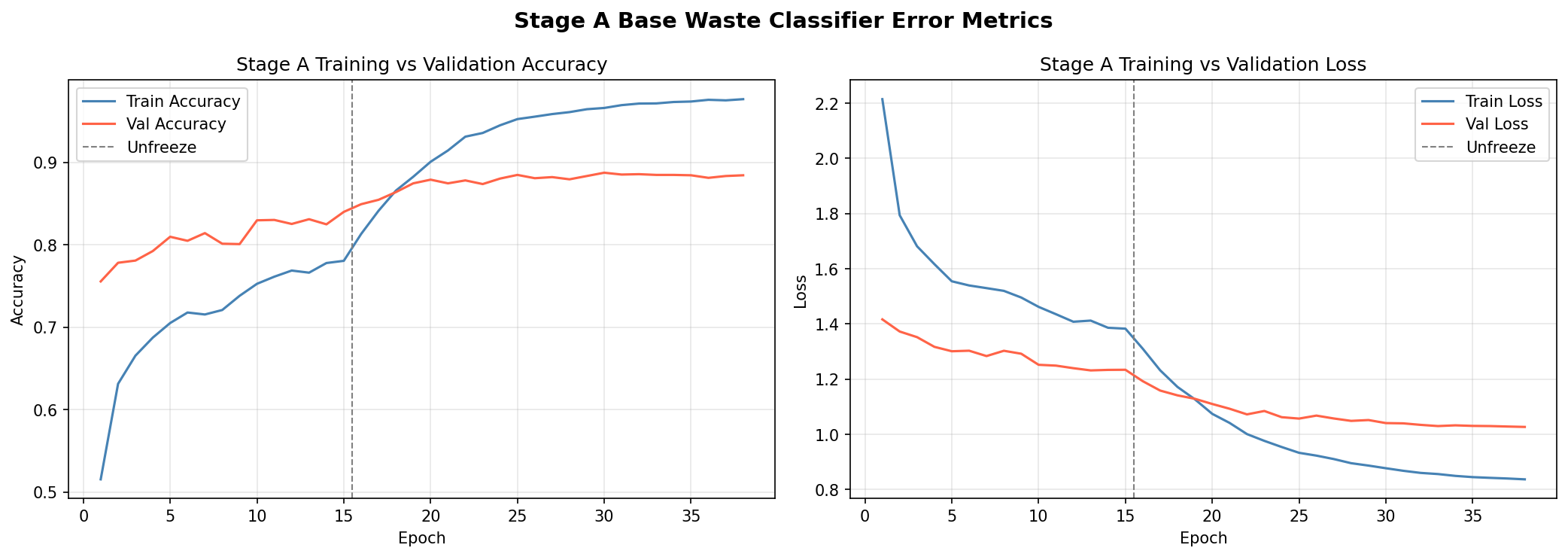}
\caption{Stage A training and validation accuracy (left) and loss (right). The dashed line marks the transition from the frozen first phase to the fine-tuning second phase, through which the curves climb without the dip that unfrozen batch normalization would otherwise produce.}
\label{fig:aerror}
\end{figure}

We begin with the training dynamics of Stage A, which Figure~\ref{fig:aerror} presents as accuracy and loss curves over the course of training, with a dashed vertical line marking the transition from the frozen first phase to the fine-tuning second phase. The curves climb smoothly through the transition rather than dipping, which confirms that holding the batch-normalization layers in inference mode achieved its intended effect.

\begin{figure}[t]
\centering
\includegraphics[width=0.98\columnwidth]{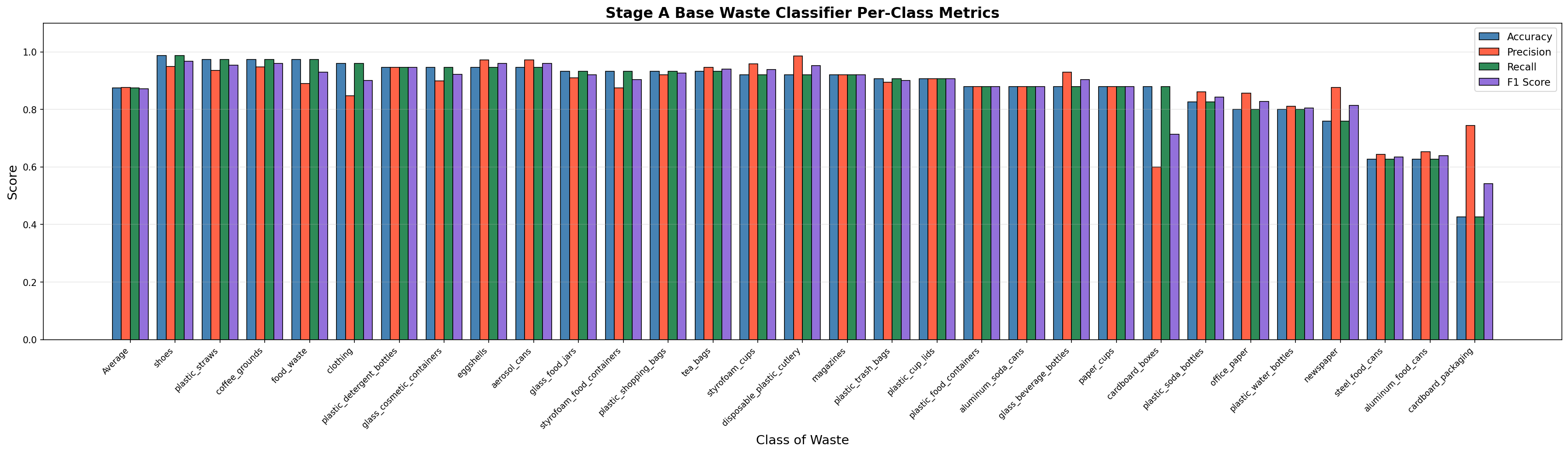}
\caption{Stage A per-class accuracy, precision, recall, and F1 score across the thirty waste classes, with the macro average shown at the left. The weakest classes correspond to visually similar materials such as the two cardboard categories.}
\label{fig:aperclass}
\end{figure}

Turning to test performance, Figure~\ref{fig:aperclass} reports the per-class accuracy, precision, recall, and F1 score for all thirty classes. Stage A achieved an average accuracy rate of 87.42 percent across all 30 classes. The weakest performers are the cardboard packaging, aluminum food cans, and steel cans. This is expected as cardboard packaging is visually similar to cardboard boxes and steel cans are visually similar to aluminum food cans.

\begin{figure}[t]
\centering
\includegraphics[width=0.98\columnwidth]{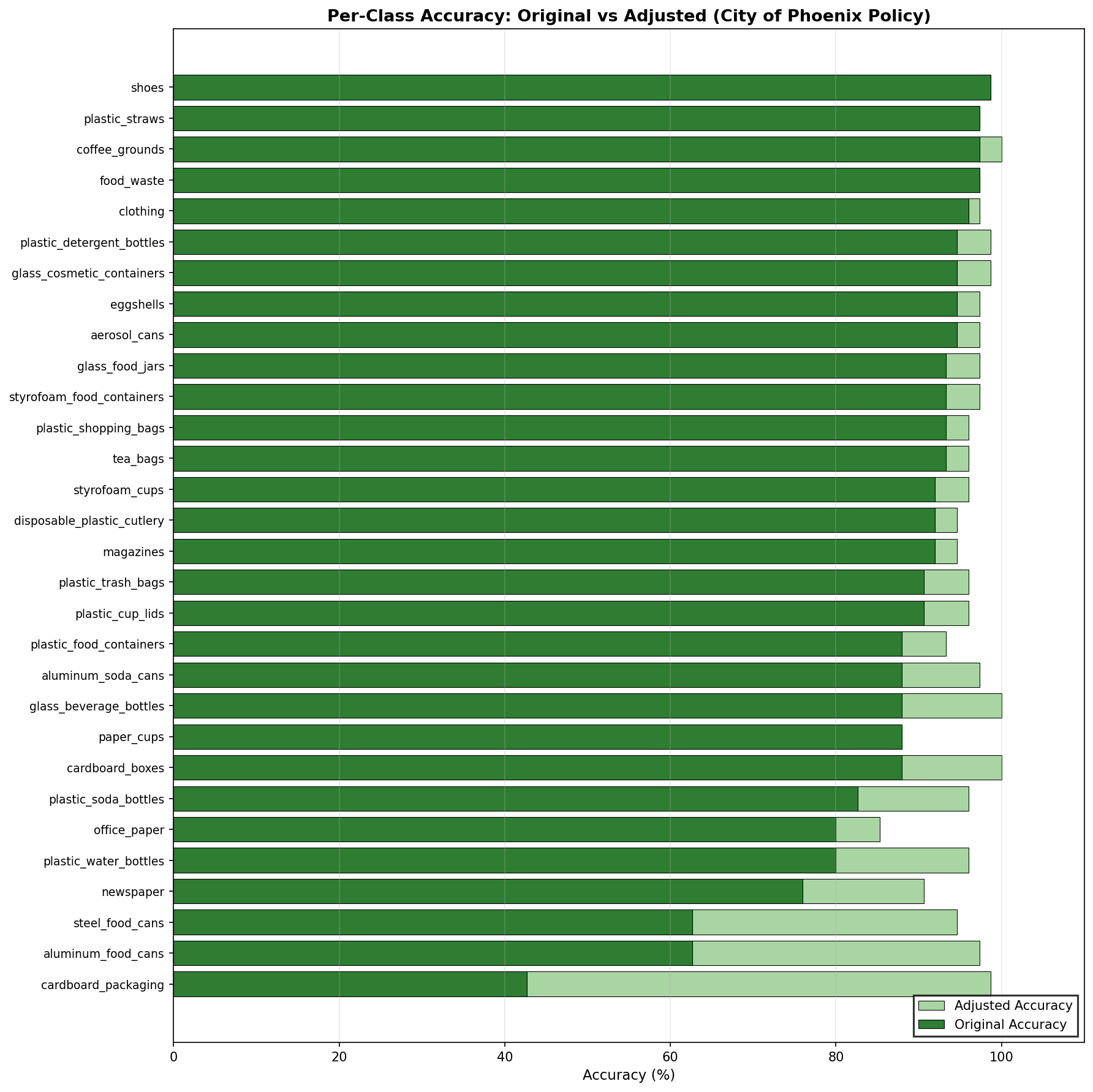}
\caption{Original versus adjusted (pathway-level) per-class accuracy for Stage A for all 30 classes.}
\label{fig:aadjusted}
\end{figure}

This observation motivates the adjusted-accuracy view in Figure~\ref{fig:aadjusted}, which counts a prediction as correct whenever the predicted class and the true class map to the same disposal pathway, even if the classes themselves differ. For example, if the model labels a cardboard-packaging item as a cardboard box, both classes still map to curbside recycling, so the user arrives at the correct bin. The adjusted accuracy is 96.13 percent, which is substantially higher than the raw class accuracy of 87.42 percent.

\begin{figure}[t]
\centering
\includegraphics[width=0.9\columnwidth]{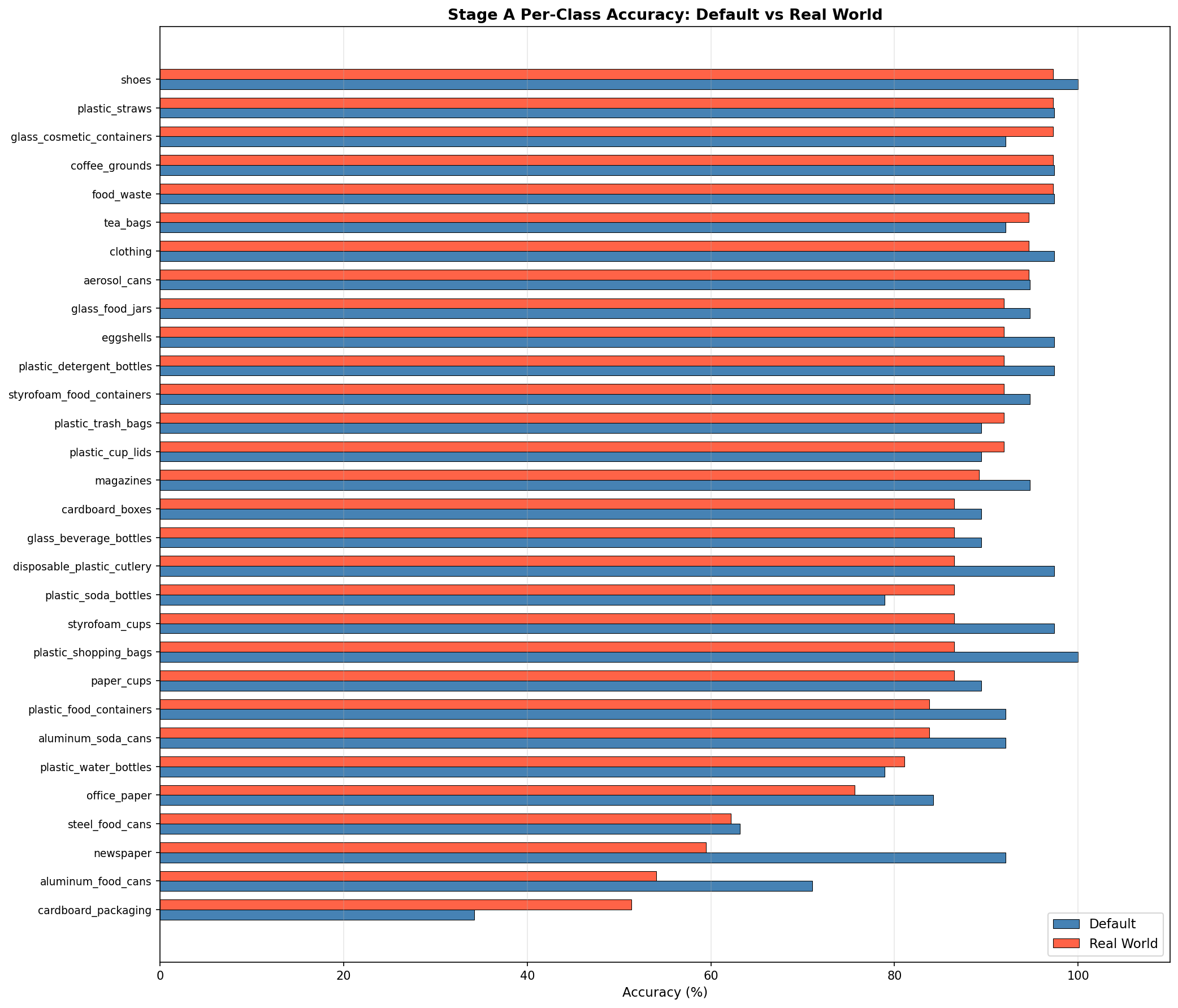}
\caption{Per-class accuracy of Stage A on the default versus the real-world test subset. The real-world subset is the more honest predictor of deployed performance because the dashboard operates on in-the-wild photographs.}
\label{fig:adomain}
\end{figure}

Finally, we compare the per-class accuracy on the default and real-world test subsets. The accuracy rate for default images is 89.21 percent and the accuracy rate for real-world images is 85.59 percent. Figure~\ref{fig:adomain} shows a visible gap between the two subsets for several classes, and this is expected because real-world photographs contain occlusion and inconsistent lighting that the model finds harder to classify.

\subsection{Stage B Results}
\begin{figure}[t]
\centering
\includegraphics[width=0.98\columnwidth]{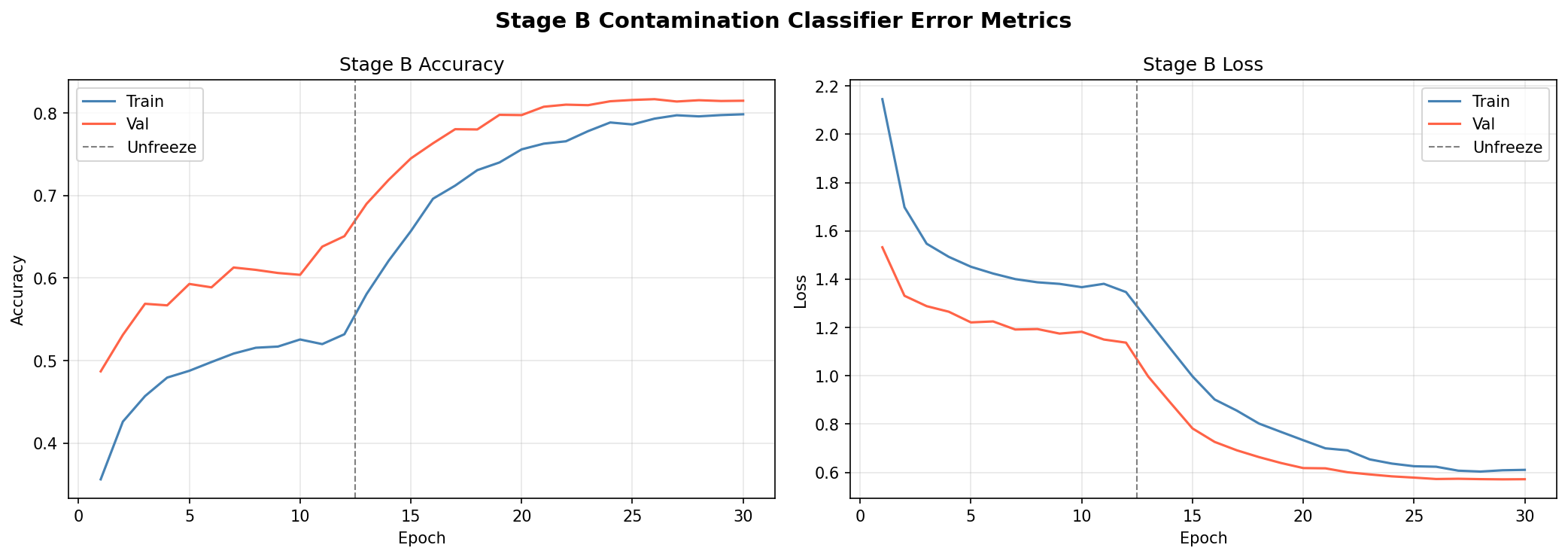}
\caption{Stage B training and validation accuracy and loss for the nine-class contamination head. The dashed line marks the phase transition.}
\label{fig:berror}
\end{figure}

The training dynamics of Stage B appear in Figure~\ref{fig:berror}, and as with Stage A the accuracy and loss curves progress through the phase transition without instability, which indicates that the contamination head converged on the synthetic data without severe overfitting to the textures.

\begin{figure}[t]
\centering
\includegraphics[width=0.98\columnwidth]{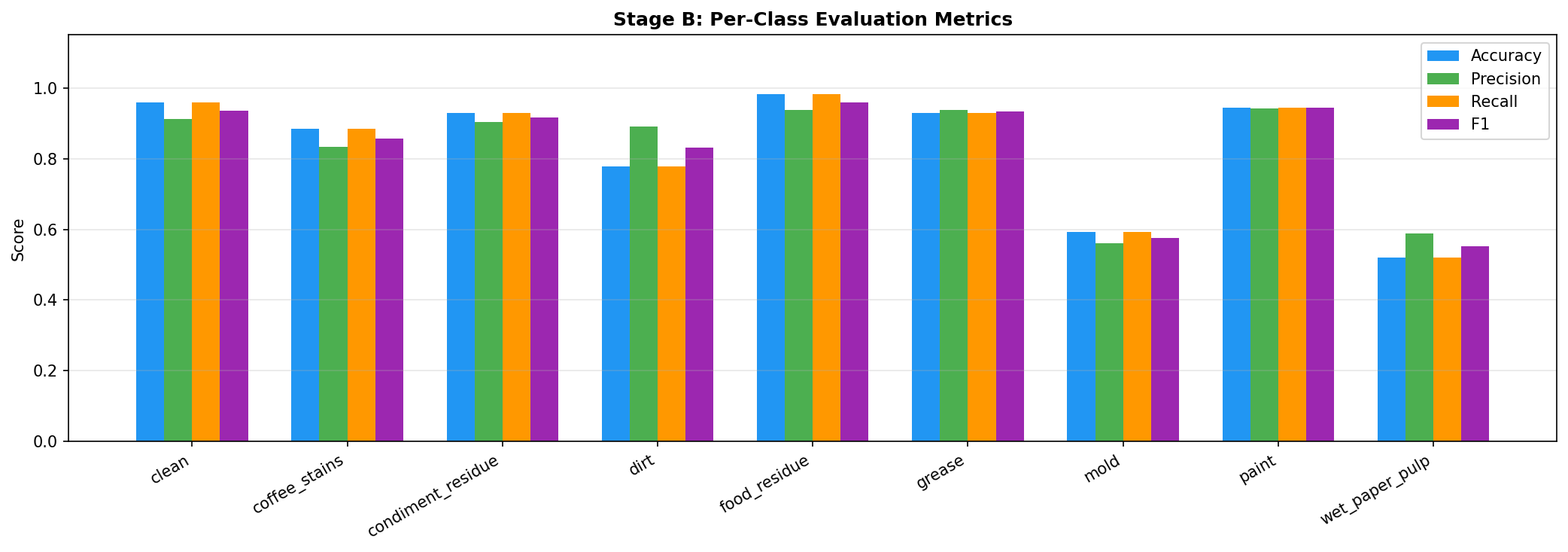}
\caption{Stage B per-class accuracy, precision, recall, and F1 across the clean class and the eight contaminant types. The mold and wet-paper-pulp classes are the most difficult because their textures are visually subtle.}
\label{fig:bperclass}
\end{figure}

The per-class results in Figure~\ref{fig:bperclass} show that Stage B distinguishes the clean class and the more visually conspicuous contaminant types, such as food residue, paint, condiment residue, and grease, with high accuracy, while two contaminant types, namely mold and wet paper pulp, are markedly more difficult, with accuracies near 0.59 and 0.52 respectively. We attribute this difficulty to the subtlety of those textures, because mold and wet pulp alter the appearance of a surface far less dramatically than a bright paint smear or a dark grease stain.

\begin{figure}[t]
\centering
\includegraphics[width=0.9\columnwidth]{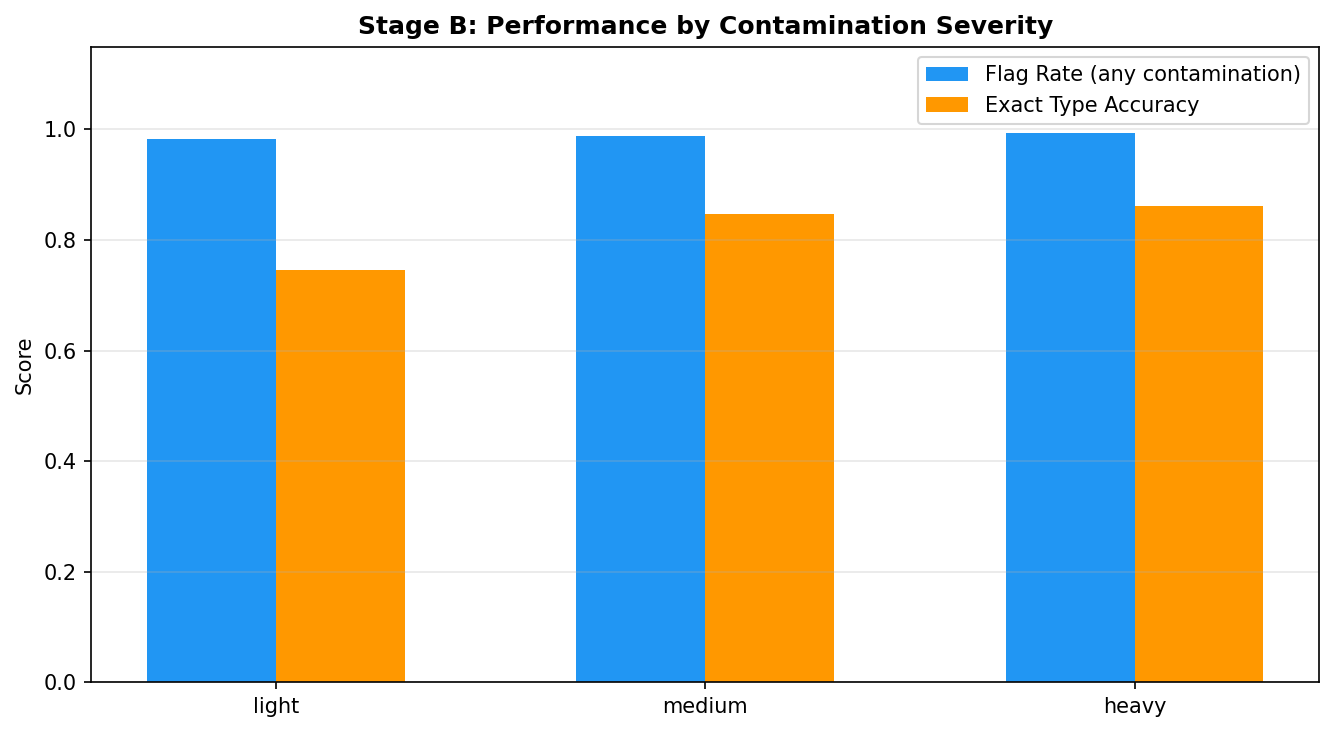}
\caption{Stage B performance by contamination severity. The flag rate is high and stable across light, medium, and heavy contamination, while exact-type accuracy increases with severity.}
\label{fig:blevel}
\end{figure}

Finally, Figure~\ref{fig:blevel} reports performance broken down by contamination severity. The flag rate, which measures whether the model predicts any non-clean class, is approximately 0.98 across all three severity levels, which means that the pipeline reliably detects the presence of contamination regardless of how light it is. The exact-type accuracy, which additionally requires the model to name the correct contaminant, rises from roughly 0.75 for light contamination to roughly 0.85 for medium and 0.86 for heavy contamination, which matches the intuition that heavier contamination presents more visible evidence for the model to work with.

\subsection{McNemar's Statistical Test}
The preceding results characterize each stage in isolation, but the purpose of the complete pipeline is to evaluate how the two stages work when combined. Specifically, the purpose is to correct a specific failure of Stage A, namely its inability to detect contamination on an item whose material it has correctly identified, and to test whether Stage B, which was trained on synthetic data, can achieve the same results on real-world objects. We assembled a test set of twenty-five real photographs of contaminated recyclable items, for which the correct disposal pathway in every case is garbage, because the contamination renders these otherwise recyclable items unsuitable for recycling.

Because the two pipelines are evaluated on the same images, their predictions are paired, and the appropriate comparison is a McNemar's test, which examines the cases on which the two pipelines disagree [19], [20]. Table~\ref{tab:mcnemar} presents the resulting two-by-two contingency table. The discordant cells are entirely one-sided, since the full pipeline was correct and Stage A was incorrect on 23 of the 25 images. Stage A alone routed only 4.0 percent of the contaminated recyclables to the garbage correctly, whereas the full pipeline routed 96.0 percent of them correctly.

\begin{table}[t]
\centering
\footnotesize
\begin{tabular}{@{}lcc@{}}
\toprule
 & \textbf{Full correct} & \textbf{Full wrong} \\
\midrule
\textbf{Stage A correct} & 1 & 0 \\
\textbf{Stage A wrong} & 23 & 1 \\
\bottomrule
\end{tabular}
\caption{Contingency table for the contaminated-recyclable test set ($n = 25$). All discordant pairs favor the full pipeline.}
\label{tab:mcnemar}
\end{table}

Because the number of discordant pairs is small, we use the exact binomial form of the test rather than the chi-squared approximation, and it yields a p-value of approximately $2.4\times10^{-7}$, which is far below the significance threshold of 0.05. We therefore conclude that the improvement contributed by the contamination stage is statistically significant on this test set, and more importantly that the improvement is concentrated precisely on the contamination cases that Stage A is structurally unable to handle, which is why we implemented Stage B.

\section{Limitations and Future Work}
\subsection{Limitations}
Several limitations should be considered alongside these results. The most significant is that Stage B is trained on synthetic contamination data. Although the synthesis pipeline produces visually plausible composites, synthetic contamination cannot perfectly reproduce the appearance of real contamination. Another limitation is that the dataset used to train Stage A is limited to thirty classes and so it does not cover every item a household might encounter. Finally, we also note that the pipeline currently has no hazardous-waste pathway, simply because no class in the dataset required one, and that the McNemar evaluation, while decisive, was conducted on a small set of twenty-five images.

\subsection{Future Work}
The limitations above suggest a clear path for future work. The highest priority is to collect a dataset of real contaminated recyclables to train Stage B to identify contamination among actual recyclable waste items. On the modeling side, moving from whole-image classification toward object detection would allow the system to handle photographs that contain several items at once, which is closer to how waste is actually encountered, and compressing the models for on-device or edge inference would reduce the dependence on a network connection and improve responsiveness. Finally, expanding the taxonomy of contaminant types and waste classes would strengthen the results of both Stage A and Stage B.

In conclusion, we have presented EcoBin, a two-stage pipeline that first identifies the disposal pathway of a waste item and then overrides recycling decisions when contamination is detected, and we have shown that the contamination stage resolves an entire category of error that a conventional classifier cannot address. By organizing the system around disposal pathways, synthetically generating the contamination dataset to train Stage B, and by evaluating the complete pipeline for statistical significance, we have connected a technical contribution to the behavioral problem at the heart of low recycling rates. We hope that the design will inform future contamination-aware sorting systems.

\begin{numrefs}
\item U.S. Environmental Protection Agency. (n.d.). \textit{National overview: Facts and figures on materials, wastes and recycling}. Retrieved from https://www.epa.gov/facts-and-figures-about-materials-waste-and-recycling/national-overview-facts-and-figures-materials
\item Cook, E., Ionkova, K., Bhada-Tata, P., Yadav, S., \& van Woerden, F. (2026). \textit{What a waste 3.0: Global snapshot of solid waste management toward circularity until 2050}. Urban Development Series. Washington, DC: World Bank. https://doi.org/10.1596/978-1-4648-2309-1
\item The Recycling Partnership. (2024). \textit{2024 state of recycling report: The present and future of residential recycling in the U.S.} Retrieved from https://recyclingpartnership.org/residential-recycling-report/
\item Fotovvatikhah, F., Ahmedy, I., Md Noor, R., \& Munir, M. U. (2025). A systematic review of AI-based techniques for automated waste classification. \textit{Sensors, 25}(10), 3181. https://doi.org/10.3390/s25103181
\item Thung, G., \& Yang, M. (2016). \textit{Classification of trash for recyclability status} [CS229 project report]. Stanford University. https://cs229.stanford.edu/proj2016/report/ThungYang-ClassificationOfTrashForRecyclabilityStatus-report.pdf
\item Proença, P. F., \& Simões, P. (2020). \textit{TACO: Trash annotations in context for litter detection} (arXiv:2003.06975) [Preprint]. arXiv. https://arxiv.org/abs/2003.06975
\item Kumsetty, N. V., Nekkare, A. B., Kamath, S. S., \& Kumar, M. A. (2022). TrashBox: Trash detection and classification using quantum transfer learning. In \textit{2022 31st Conference of Open Innovations Association (FRUCT)} (pp. 125--130). IEEE. https://doi.org/10.23919/FRUCT54823.2022.9770922
\item Ibrahim, K., Savage, D. A., Schnirel, A., Intrevado, P., \& Interian, Y. (2019). \textit{ContamiNet: Detecting contamination in municipal solid waste} (arXiv:1911.04583) [Preprint]. arXiv. https://arxiv.org/abs/1911.04583
\item Pan, S. J., \& Yang, Q. (2010). A survey on transfer learning. \textit{IEEE Transactions on Knowledge and Data Engineering, 22}(10), 1345--1359. https://doi.org/10.1109/TKDE.2009.191
\item Deng, J., Dong, W., Socher, R., Li, L.-J., Li, K., \& Fei-Fei, L. (2009). ImageNet: A large-scale hierarchical image database. In \textit{2009 IEEE Conference on Computer Vision and Pattern Recognition} (pp. 248--255). IEEE. https://doi.org/10.1109/CVPR.2009.5206848
\item Tan, M., \& Le, Q. V. (2021). EfficientNetV2: Smaller models and faster training. In \textit{Proceedings of the 38th International Conference on Machine Learning} (PMLR 139, pp. 10096--10106).
\item Shorten, C., \& Khoshgoftaar, T. M. (2019). A survey on image data augmentation for deep learning. \textit{Journal of Big Data, 6}, 60. https://doi.org/10.1186/s40537-019-0197-0
\item Ioffe, S., \& Szegedy, C. (2015). Batch normalization: Accelerating deep network training by reducing internal covariate shift. In \textit{Proceedings of the 32nd International Conference on Machine Learning} (PMLR 37, pp. 448--456).
\item Loshchilov, I., \& Hutter, F. (2017). SGDR: Stochastic gradient descent with warm restarts. In \textit{5th International Conference on Learning Representations (ICLR)}. https://arxiv.org/abs/1608.03983
\item Kingma, D. P., \& Ba, J. (2015). Adam: A method for stochastic optimization. In \textit{3rd International Conference on Learning Representations (ICLR)}. https://arxiv.org/abs/1412.6980
\item King, A. (n.d.). \textit{Recyclable and household waste classification dataset} [Data set]. Kaggle. Retrieved from https://www.kaggle.com/datasets/alistairking/recyclable-and-household-waste-classification
\item Srivastava, N., Hinton, G., Krizhevsky, A., Sutskever, I., \& Salakhutdinov, R. (2014). Dropout: A simple way to prevent neural networks from overfitting. \textit{Journal of Machine Learning Research, 15}(1), 1929--1958.
\item Qin, X., Zhang, Z., Huang, C., Dehghan, M., Zaïane, O. R., \& Jagersand, M. (2020). U2-Net: Going deeper with nested U-structure for salient object detection. \textit{Pattern Recognition, 106}, 107404. https://doi.org/10.1016/j.patcog.2020.107404
\item McNemar, Q. (1947). Note on the sampling error of the difference between correlated proportions or percentages. \textit{Psychometrika, 12}(2), 153--157. https://doi.org/10.1007/BF02295996
\item Dietterich, T. G. (1998). Approximate statistical tests for comparing supervised classification learning algorithms. \textit{Neural Computation, 10}(7), 1895--1923. https://doi.org/10.1162/089976698300017197
\end{numrefs}

\end{document}